%% file: main.tex
\documentclass[10pt,twocolumn,letterpaper]{article}

\usepackage[pagenumbers]{cvpr} %

\usepackage{booktabs}
\usepackage{multirow}
\usepackage{colortbl}
\usepackage{graphicx}
\usepackage{xcolor}
\usepackage[percent]{overpic}
\usepackage[usestackEOL]{stackengine}
\usepackage{caption}
\usepackage{subcaption}

\usepackage[accsupp]{axessibility}  %

\input{preamble}
\definecolor{cvprblue}{rgb}{0.21,0.49,0.74}
\usepackage[pagebackref,breaklinks,colorlinks,allcolors=cvprblue]{hyperref}
\usepackage{tikz}
\usepackage{pgfplots}
\pgfplotsset{compat=1.18} 

\def\paperID{6537} %
\def\confName{CVPR}
\def\confYear{2025}

\newcommand{\prada}{PRaDA}

\title{\prada: Projective Radial Distortion Averaging}

\author{Daniil Sinitsyn$^{1,2}$, Linus Härenstam-Nielsen$^{1,2}$,  Daniel Cremers$^{1,2}$\\
$^1$Technical University of Munich, $^2$Munich Center for Machine Learning\\
{\tt\small daniil.sinitsyn@tum.de, linus.nielsen@tum.de, cremers@tum.de}
}

\newcommand \eq[1]{\begin{equation}\begin{aligned}#1\end{aligned}\end{equation}}

\newcommand{\pfirst}[1]{p_{#1}}
\newcommand{\psecond}[1]{q_{#1}}
\newcommand{\upfirst}[1]{p_{#1}^u}
\newcommand{\upsecond}[1]{q_{#1}^{u}}
\newcommand{\upfirstbar}[1]{\bar{p}^{u}_{#1}}
\newcommand{\upsecondbar}[1]{\bar{q}^{u}_{#1}}
\newcommand{\U}[1]{U_{#1}}
\newcommand{\param}{\theta}
\newcommand*\diff{\mathop{}\!\mathrm{d}}
\newcommand*{\inparagraph}[1]{\noindent\textbf{#1}}

\begin{document}
\maketitle

\begin{abstract}
We tackle the problem of automatic calibration of radially distorted cameras in challenging conditions.
Accurately determining distortion parameters typically requires either 1) solving the full Structure from Motion (SfM) problem involving camera poses, 3D points, and the distortion parameters, which is only possible if many images with sufficient overlap are provided, or 2) relying heavily on learning-based methods that are comparatively less accurate.
In this work, we demonstrate that distortion calibration can be decoupled from 3D reconstruction, maintaining the accuracy of SfM-based methods while avoiding many of the associated complexities. 
This is achieved by working in Projective space, where the geometry is unique up to a homography, which encapsulates all camera parameters except for distortion.
Our proposed method, Projective Radial Distortion Averaging, averages multiple distortion estimates in a fully projective framework without creating 3d points and full bundle adjustment. 
By relying on pairwise projective relations, our methods support any feature-matching approaches without constructing point tracks across multiple images.

\end{abstract}

\input{sec/1_introduction}
\input{sec/2_related_work}

\input{sec/3_method}

\input{sec/4_implementation}
\begin{table*}[t]
\centering
\footnotesize
\begin{tabular}{lccc|ccc|ccc|ccc|ccc}
\toprule
& \multicolumn{3}{c}{ScanNet++} & \multicolumn{3}{c}{ETH3D (cam 4)} & \multicolumn{3}{c}{ETH3D (cam 5)} & \multicolumn{3}{c}{KITTI-360 (cam 2)} & \multicolumn{3}{c}{KITTI-360 (cam 3)} \\
\cmidrule(lr){2-4} \cmidrule(lr){5-7} \cmidrule(lr){8-10} \cmidrule(lr){11-13} \cmidrule(lr){14-16}
Method & Min & Mean & Max & Min & Mean & Max & Min & Mean & Max & Min & Mean & Max & Min & Mean & Max \\
\midrule
Colmap~\cite{schoenberger2016sfm}      & 0.7  & 2.0   & 6.4   & 13.2 & 26.0  & 44.5  & 18.6 & 25.1  & 37.1  & 117.1 & 125.5 & 136.4 & 97.6 & 112.4 & 127.4 \\
Glomap~\cite{pan2024glomap}      & 0.6  & 1.8   & 3.4   & 10.0 & 18.4  & 29.3  & 8.6 & 19.6  & 58.0  & 85.9 & 122.0 & 138.3  & 91.9 & 113.3 & 128.7 \\
DroidCalib~\cite{hagemann2023droidCalib}  & 0.4  & 1.2  & 2.6  & 19.9 & 36.3  & 58.4 & 18.0 & 46.4 & 71.1 & 98.7 & 102.2 & 105.2  & 120.9 & 128.1 & 135.3 \\
GeoCalib~\cite{veicht2024geocalib}    & 4.19  & 4.6   & 4.9   & 30.1  & 35.8   & 43.0  & 26.6  & 34.6   & 123.1 & 124.4 & 125.5  & 63.8  & 122.0 & 123.1 & 124.3 \\
DeepCalib~\cite{bogdan2018deepcalib}    & 0.8  & 10.8   & 30.7   & 18.1  & 20.9   & 23.6   & 13.7  & 18.2   & \textbf{23.5}   & 141.0 & 160.5 & 174.0  & 109.4 & 153.2 & 177.1 \\
\midrule
Ours (w. SIFT) & \textbf{0.1}  & \textbf{0.6}   & \textbf{1.7}   & \textbf{2.6}  & \textbf{5.3}  & \textbf{10.4}  & \textbf{3.2}  & \textbf{14.4}  & 28.8  & \textbf{42.3}  & \textbf{44.8}  & \textbf{46.2}  & \textbf{47.5}  & \textbf{50.2}  & \textbf{55.1} \\
\bottomrule
\end{tabular}

\caption{\textbf{Performance comparison on challenging datasets:} Minimum, mean, and maximum focal-adjusted reprojection errors (in pixels) are reported for selected baseline methods.  The reported errors represent the values for all sequences within each dataset. Our method uses the same SIFT matches as COLMAP and GLOMAP to ensure a fair comparison. For datasets with multiple camera models, errors are reported for each model independently, with the corresponding model in parentheses.}

\label{tab:projection_errors}
\end{table*}

\begin{figure}[t]
    \centering
    \begin{subfigure}[t]{0.4\columnwidth}
        \centering
        \includegraphics[width=\columnwidth]{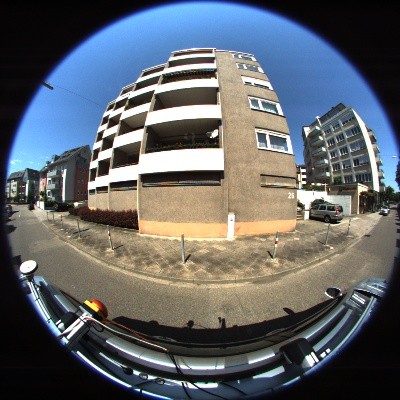}
        \phantomsubcaption
        \label{fig:kitti360}
    \end{subfigure}\quad
    \begin{subfigure}{0.4\columnwidth}
        \centering
        \includegraphics[width=\columnwidth]{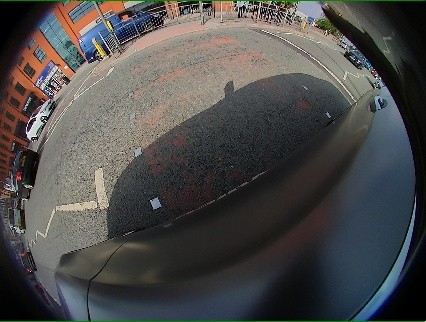}
        \phantomsubcaption
        \label{fig:woodscape}
    \end{subfigure}
    
    \vspace{-0.7em} %
    \makebox[0.4\columnwidth]{\small a) KITTI-360~\cite{Liao2022PAMI}}\hspace{0.3cm}
    \makebox[0.4\columnwidth]{\small b) WoodScape~\cite{Yogamani_2019_ICCV}}
    \vspace{-0.7em} %
    \caption{Example of images from challenging sequences.}
    \label{fig:dataset_sample}
\end{figure}

\section{Experiments}
\label{sec:experiments}

\subsection{Metrics}
\label{sec:metrics}
To measure the accuracy of the distortion estimate, we use the Reprojection Error (RE), computed as:
\eq{
    \text{RE} = \frac{1}{\lvert \Omega \rvert}\sum\limits_{p\in\Omega} \left\lVert \pi_\theta \left( \pi^{-1} \left(p \right) \right) - p \right\rVert.
    \label{eq:metric_reprojection}
}
Where $p \in \Omega$ are the image pixel coordinates, $\pi$ is the projection function of the ground truth camera model, and $\pi_\theta$ is the projection function of the estimated model with parameters $\theta$. 
MRE is widely used to compare camera models \cite{veicht2024geocalib,lochman2021minimal,bogdan2018deepcalib,Kukelova2015F10}. It provides an interpretable measure of accuracy in units of pixels in the original image.

For a fair comparison of the distortion estimation only, we compute Focal-Adjusted RE (FA-RE), which is obtained by selecting the focal length that minimizes \cref{eq:metric_reprojection}. The error thus reflects the model's ability to represent distortion uniformly across the image. This formulation helps to better capture the true error in highly distorted regions.%

We also evaluate the impact of our method on the downstream task of SfM. We measure the deviation in translation direction and rotation angle between the estimated and ground truth relative poses for each possible image pair, as in GLOMAP~\cite{pan2024glomap}.This metric is more robust to completely incorrect estimation of a subset of poses than an absolute pose error. %

\subsection{Datasets}

\inparagraph{ETH3D~\cite{schops2017eth3D}:} A multi-view stereo benchmark dataset comprising video sequences captured by camera rigs and ground-truth calibrations in COLMAP format. We use all frames from all 5 sequences from the ``Low-res many-view'' category, which includes cameras with significant distortion. 

\inparagraph{KITTI 360~\cite{Liao2022PAMI}:} An extension of the KITTI dataset that includes 360-degree panoramic images created from two 180$^\circ$ fisheye cameras. For evaluation, we use the 600 first frames for each camera from each sequence.

\inparagraph{ScanNet++~\cite{yeshwanth2023scannet++}:} This dataset provides fisheye-lens DSLR images across various indoor scenes. Each scene includes continuous sequences of 200+ images. To evaluate performance, we use all frames from the test set sequences provided by the dataset.

\inparagraph{WoodScape~\cite{Yogamani_2019_ICCV}:} WoodScape is a fisheye camera dataset comprising over 10,000 images captured by four 180$^\circ$ fisheye cameras mounted on a vehicle. The images are taken at sparse timestamps. This makes the application of feature-based methods particularly challenging. For DeepCalib and Geocalib, we use all available frames.

KITTI-360 and WoodScape are particularly challenging as the images are heavily distorted; see \cref{fig:dataset_sample}.
We estimate one camera model per distinct physical camera presented in datasets. That is: 2 cameras for ETH3D, 2 for KITTI, one for ScanNet++, and 4 for WoodScape.

\begin{figure}
    \centering
    \includegraphics[width=\columnwidth]{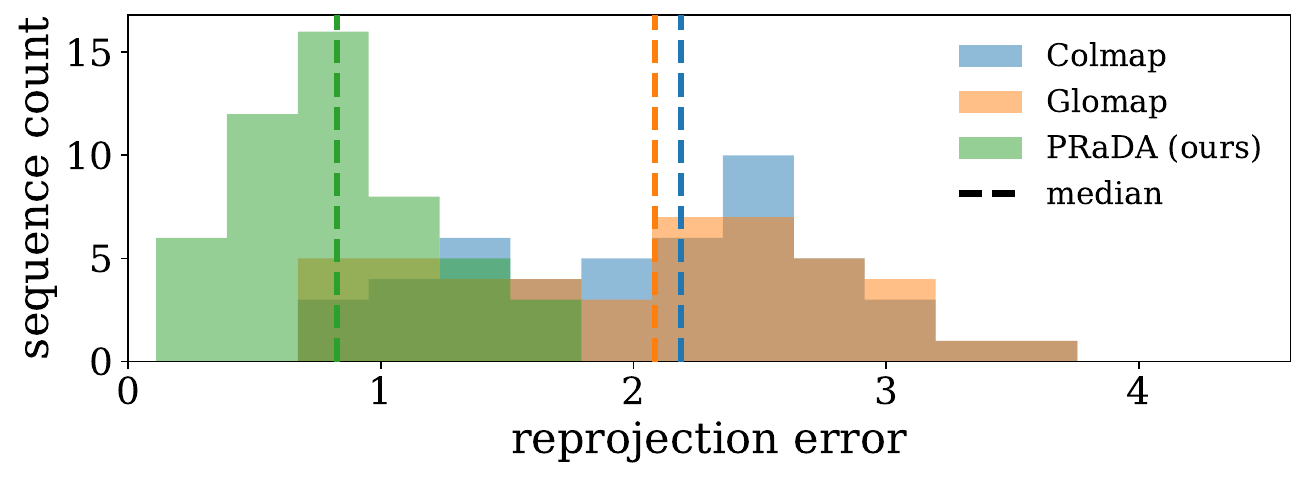}
    \vspace{-0.8cm}
    \caption{\textbf{ScanNet++ by sequence:} Our method consistently outperforms SfM-based methods Colmap~\cite{schoenberger2016mvs,schoenberger2016sfm} and Glomap~\cite{pan2024glomap} in terms of calibration accuracy. 
    }
    \label{fig:scanet_hist}
\end{figure}

\begin{figure*}
\centering
\setlength{\tabcolsep}{2pt} %
\renewcommand{\arraystretch}{1.1} %
\begin{tabular}{ccccccc}
 & Ground Truth & Ours & Geocalib & Deepcalib & COLMAP & GLOMAP \\
\midrule
\rotatebox{90}{WoodScape} & 
\includegraphics[width=0.14\linewidth]{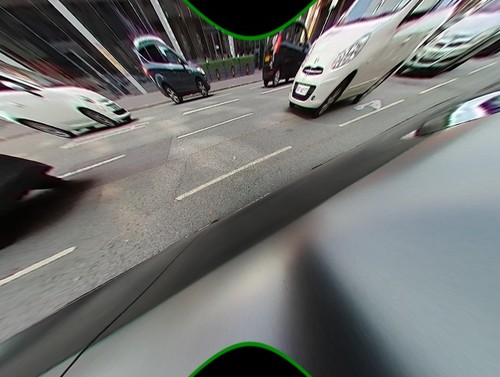} & 
\includegraphics[width=0.14\linewidth]{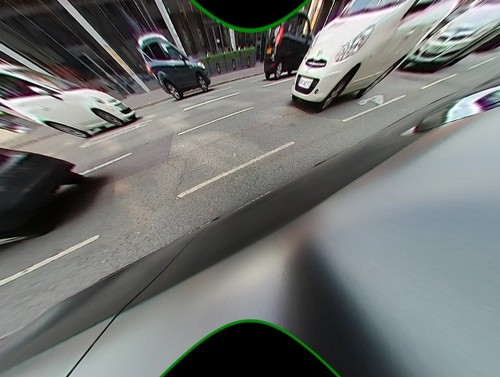} & 
\includegraphics[width=0.14\linewidth]{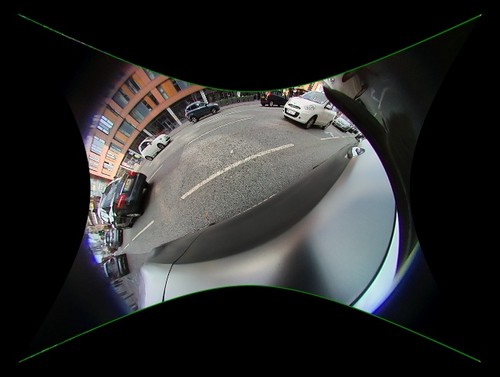} & 
\includegraphics[width=0.14\linewidth]{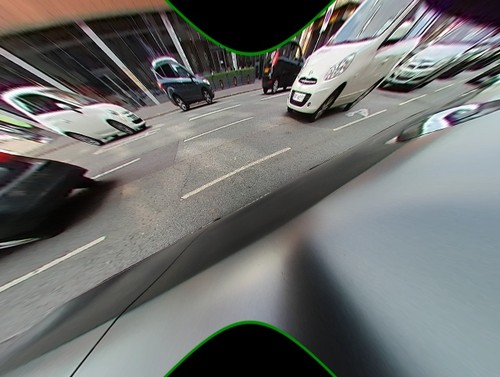} & 
\includegraphics[width=0.14\linewidth]{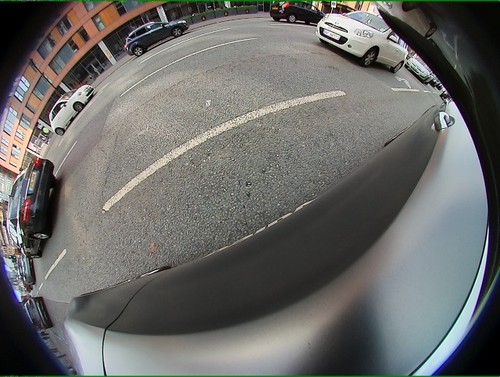} & 
\includegraphics[width=0.14\linewidth]{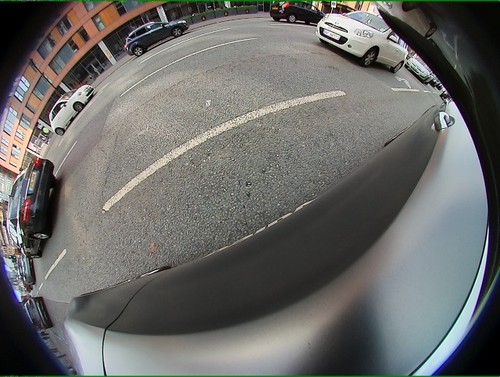} \\

\rotatebox{90}{WoodScape} & 
\includegraphics[width=0.14\linewidth]{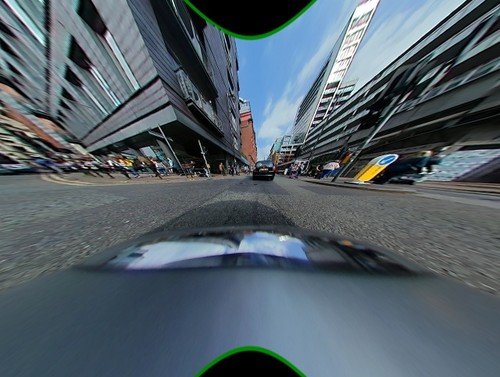} & 
\includegraphics[width=0.14\linewidth]{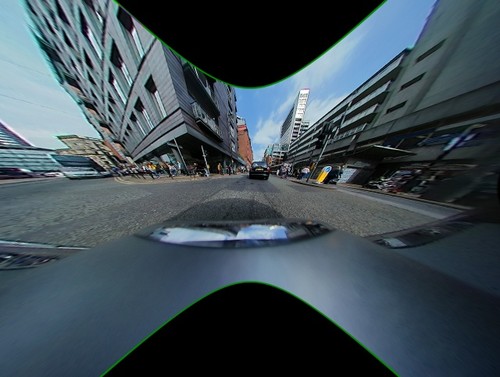} & 
\includegraphics[width=0.14\linewidth]{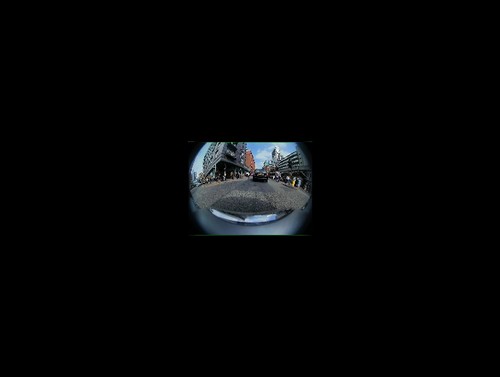} & 
\includegraphics[width=0.14\linewidth]{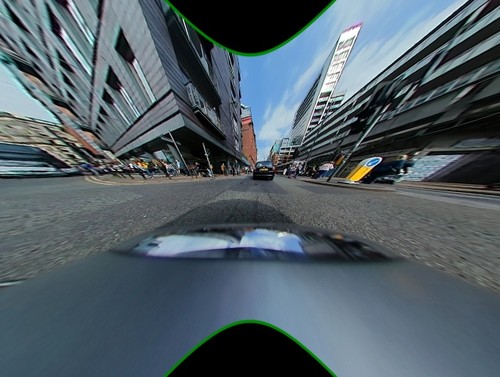} & 
\includegraphics[width=0.14\linewidth]{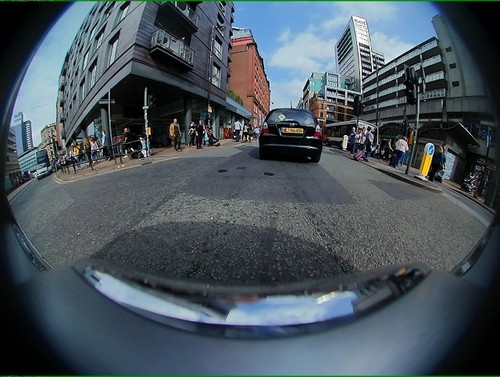} & 
\includegraphics[width=0.14\linewidth]{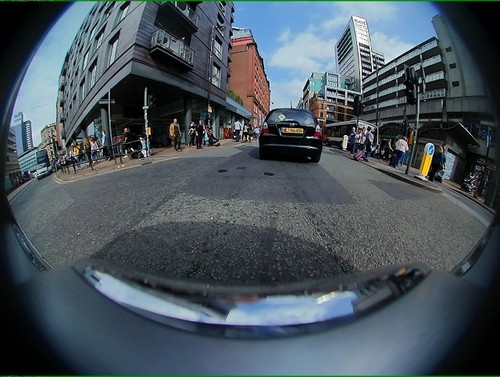} \\

\rotatebox{90}{KITTI-360 } & 
\includegraphics[width=0.14\linewidth]{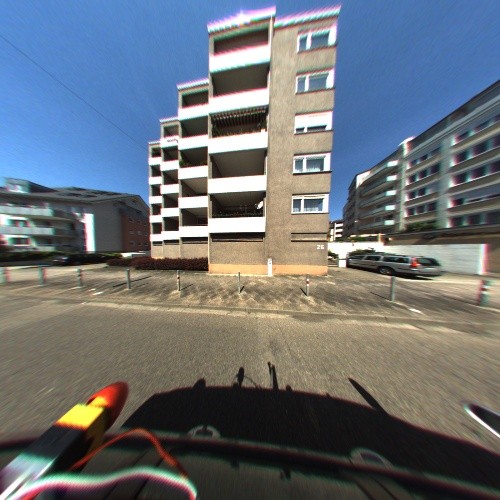} & 
\includegraphics[width=0.14\linewidth]{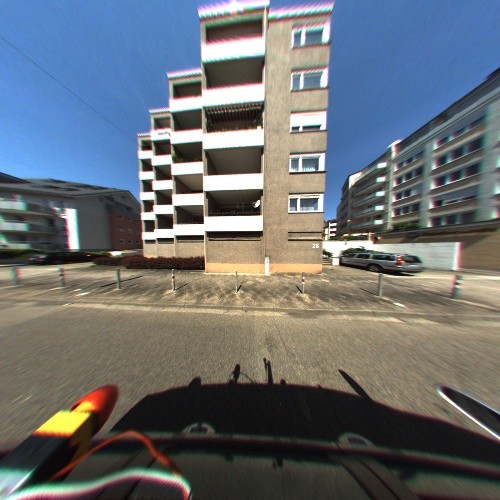} & 
\includegraphics[width=0.14\linewidth]{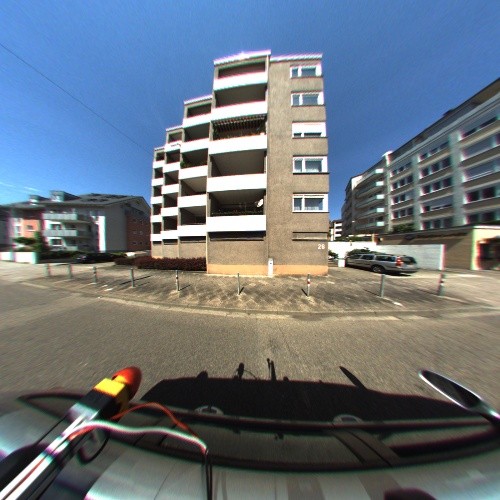} & 
\includegraphics[width=0.14\linewidth]{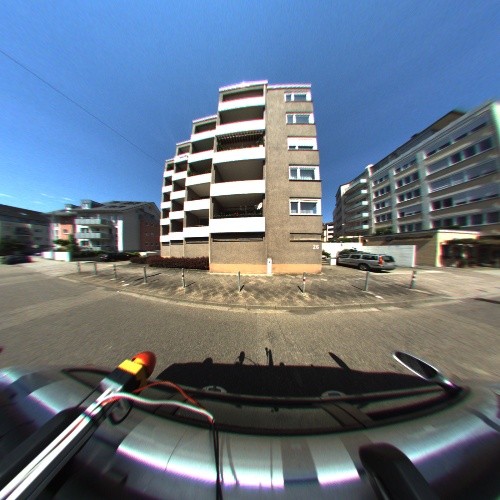} & 
\includegraphics[width=0.14\linewidth]{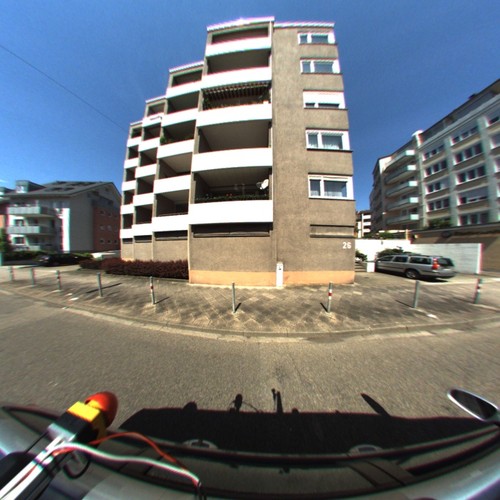} & 
\includegraphics[width=0.14\linewidth]{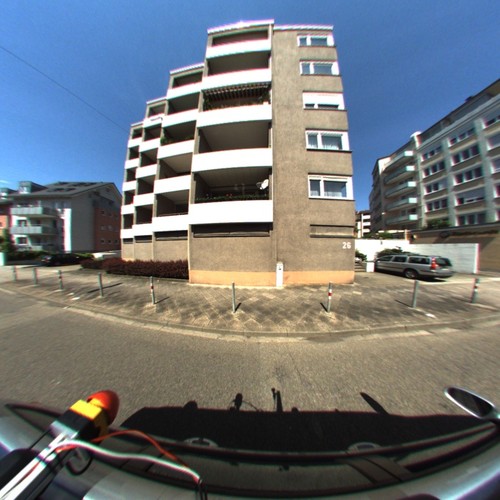} \\

\rotatebox{90}{ScanNet++ } & 
\includegraphics[width=0.14\linewidth]{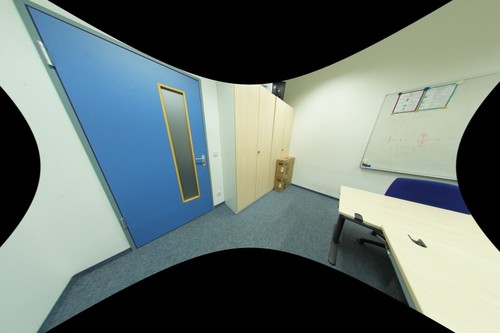} & 
\includegraphics[width=0.14\linewidth]{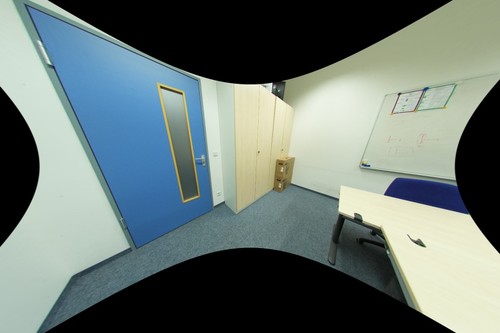} & 
\includegraphics[width=0.14\linewidth]{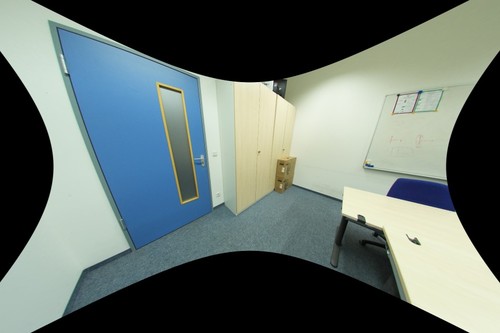} & 
\includegraphics[width=0.14\linewidth]{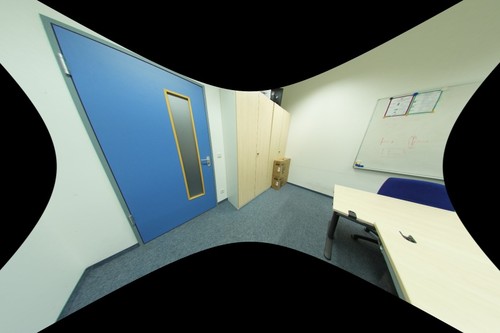} & 
\includegraphics[width=0.14\linewidth]{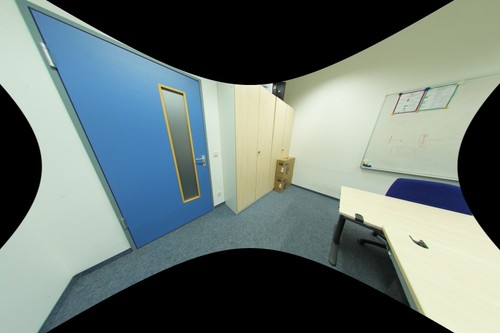} & 
\includegraphics[width=0.14\linewidth]{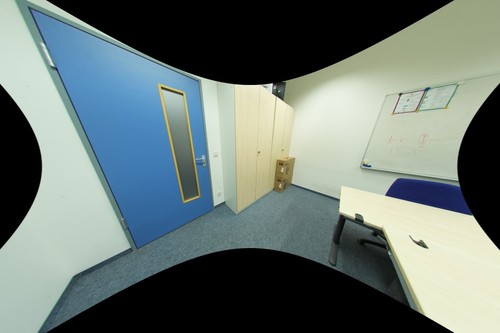} \\
\end{tabular}

\caption{\textbf{Qualitative results:} Comparison of estimated camera models on challenging datasets. The models are adjusted so that the center of an image approximately matches the center of the image undistorted by the ground truth model.}
\label{fig:qualitative_results}
\end{figure*}

\subsection{Results}

\paragraph{Baselines:} We compare our method to several common and state-of-the-art autocalibration approaches: \textbf{COLMAP~\cite{schoenberger2016sfm, schoenberger2016mvs}}, which incrementally estimates the camera model using a dense subset of images; \textbf{GLOMAP~\cite{pan2024glomap}}, which uses global optimization in SfM similar to our approach; \textbf{DroidCalib~\cite{hagemann2023droidCalib}}, a SLAM-based method that optimizes camera parameters from video input; \textbf{GeoCalib~\cite{veicht2024geocalib}}, which jointly refines the camera model and gravity direction from a single image; and \textbf{DeepCalib~\cite{bogdan2018deepcalib}}, which directly regresses camera intrinsic and distortion parameters using Mei's camera model~\cite{mei2007single}.
 Additional implementation details are given in supplementary \cref{sec:supp_implementation_details}.

\begin{table}[t]
\centering
\footnotesize
\begin{tabular}{lcccc}
\toprule
Method &  FV &  RV &  MVR &  MVL \\
\midrule
Ours      &  51.2 & 22.2 & \textbf{7.7} & \textbf{9.4} \\
DeepCalib &  \textbf{9.7} & \textbf{14.5} & 14.5 & 21.5 \\
GeoCalib  &  98.0 &96.2 & 93.0 & 93.9 \\
\bottomrule
\end{tabular}
\caption{\textbf{WoodScape dataset:} The mean focal-adjusted reprojection error on the WoodScape dataset is reported for each camera as follows: \textbf{FV} points forward, \textbf{RV} points backward, \textbf{MVR} represents the right-side camera, and \textbf{MVL} represents the left-side camera.}
\label{tab:woodscape_results}
\end{table}

\inparagraph{Quantitative Results:} The results of the reprojection errors are summarized in \cref{tab:projection_errors}. 
To ensure fairness in the comparison, we run our solution on the same SIFT matches as COLMAP and GLOMAP. If either COLMAP or GLOMAP splits the 3D scene into multiple models, we choose the one with more registered images. 
We run DroidCalib several times, varying parameters such as image step and number of frames. Geocalib and DeepCalib are run on all images. For DroidCalib, Deepcalib, and Geocalib, we report \textit{the mean error} over all models for each sequence.

Our proposed solution consistently outperforms all baseline methods across all datasets, achieving lower minimum reprojection errors. Despite these improvements, due to the radially symmetric distortion assumption, the proposed method cannot achieve a subpixel reprojection error on datasets with ground-truth models that are not radially symmetric, such as ETH3D \cite{schops2017eth3D} or KITTI-360\cite{Liao2022PAMI}.

\cref{fig:scanet_hist} shows the histogram of reprojection errors per sequence on the test split of ScanNet++. Our method demonstrates a significantly higher concentration of errors below 1 pixel, outperforming both COLMAP and GLOMAP in this critical accuracy range.

The results on WoodScape are summarized in \cref{tab:woodscape_results}. We compare our solution to DeepCalib and GeoCalib because other methods do not produce meaningful results. Our solution was applied to the first 60 frames of each camera. Increasing the number of frames had no significant effect on the results. We used LOFTR~\cite{sun2021loftr} to build point correspondences. DeepCalib and GeoCalib used all available images for each camera. We report the mean error of all estimated models.
As expected, we get worse results for cameras pointing backward and forward. This means that the center of the distortion coincides with or close to the projection of the second camera onto the image, which corresponds to the degenerative case for distortion estimation from correspondences~\cite{henrique2013radial, wu2014critical}. Intuitively, this is the case when all epipolar lines are straight, regardless of camera distortion.

We compare the accuracy of GLOMAP with and without the distortion model initialization from PRaDA in \cref{fig:glomap_init_exp}.
We evaluate angular errors for relative poses on the Sparse test images from the test split of ScanNet++ \cite{yeshwanth2023scannet++},  which is a challenging dataset containing extreme baselines with low overlap. 
The results demonstrate the benefits of a decoupled calibration stage.
In contrast, joint estimation of camera calibration and 3D geometry often leads to poor reconstruction performance in non-ideal settings.

\inparagraph{Qualitative results:} \cref{fig:qualitative_results} presents qualitative results, showing the visual undistortion of an image from multiple datasets. If the method does not work on the dataset, we show the original image, such as for COLMAP and GLOMAP on WoodScape. 
Since we are not estimating the focal length, we match it to the one of the ground truth model. To do it, we rely on the observation that a distorted image can be approximated by a pinhole camera near its center. Specifically, for a given focal length $f$ and a small pixel displacement $dx$, the angular resolution near the image center, denoted by $\theta$, can be described as:
\eq{
\frac{dx}{f} \approx \tan(\theta)
}
For small angles $\tan(\theta) \approx \theta$. Then for $f$:
\eq{
f \approx \frac{dx}{\theta}
}
The angle $\theta$ is defined as the angle between the optical ray corresponding to the principal point and the ray corresponding to the back projection of the displacement $dx$.

This ensures that all models behave similarly near the center of the image, making it possible to undistort the images defining the same FOV.

\begin{table}
    \centering    
    \begin{tabular}{lcc}
        \toprule
        & PRaDA + GLOMAP & GLOMAP \\
        \midrule
        $r_{err}$ (deg) & \textbf{0.18/4.51/44.56} & 0.25/28.99/118.6 \\
        $t_{err}$ (deg) & \textbf{0.26/8.70/81.07} & 0.33/27.39/95.76 \\
        \bottomrule
    \end{tabular}
    \caption{\textbf{Sparse ScanNet with different initialization:} Min/Mean/Max angular errors of relative poses for GLOMAP and GLOMAP initialized with PRaDA. These results highlight the benefits of PRaDA for 3D reconstruction.}
    \label{fig:glomap_init_exp}
\end{table}
\section{Discussion}

Similar to COLMAP~\cite{schoenberger2016sfm,schoenberger2016mvs} and GLOMAP~\cite{pan2024glomap}, our method relies on the performance of the matcher. If the matcher has been trained or optimized primarily for pinhole settings, its errors can noticeably impact our results. This is because we use the Levenberg-Marquardt algorithm as a nonlinear optimizer, which assumes normally distributed errors. This condition may not hold in distorted regions, resulting in reduced accuracy. This can be properly modeled by estimating the error distribution. 

Like COLMAP and GLOMAP, which use predefined thresholds for RANSAC and inlier estimation, we also rely on them. We would like to point out that it is possible to overcome the thresholds completely. Using the $\sigma$ or $\sigma\text{++}$ consensus introduced by \citet{barath2019magsac, barath2020magsac++} through all steps of the proposed algorithm makes this possible. However, its application requires thoughtful design.  We see this as a promising direction for future research.

\section{Conclusion}

This paper proposes a new method for estimating radial distortion in a projective setting. Our approach integrates distortion modeling to each stage of the global projective reconstruction pipeline, with model averaging based on pixel errors in the image plane relative to less expressive models. This careful design allows for more accurate distortion correction.
We demonstrate that our method outperforms modern techniques, achieving the lowest error on almost all datasets. We show that it can estimate extremely distorted images with a $180^\circ$ lens.

KITTI-360~\cite{Liao2022PAMI} and WoodScape~\cite{Yogamani_2019_ICCV} serve as strong examples of the robustness of the proposed solution, as they use cameras with an exact 180$^\circ$ field of view. This demonstrates the ability of the proposed solution to effectively handle extreme wide-angle fisheye cameras.

\section*{Acknowledgments}
This work was supported by the ERC Advanced Grant ``SIMULACRON'' (agreement \#884679), the GNI Project ``AI4Twinning'', and the DFG project CR 250/26-1 ``4D YouTube''.

{
    \small
    \bibliographystyle{ieeenat_fullname}
    \bibliography{main}
}

\input{sec/X_suppl}

\end{document}

%% file: sec/1_introduction.tex
\section{Introduction}
\label{sec:introduction}

Having access to an accurate camera model is foundational to essentially all geometric computer vision algorithms, including Structure from Motion (SfM), Simultaneous Localization and Mapping (SLAM), and novel-view synthesis (NVS).
However, in many cases, prior accurate intrinsics are not available.
For instance, if images are sourced from the internet or if camera parameters have drifted over time due to wear and tear.
In such cases, intrinsics must instead be estimated using geometric principles or learned from data.
Despite active research, accurately determining intrinsics remains a challenging problem with no established go-to method.
Methods based on full SfM and bundle adjustment~\cite{schoenberger2016mvs,schoenberger2016sfm} often fail to converge without a good initialization, and learning-based methods~\cite{veicht2024geocalib} tend to lack accuracy and robustness.

In this work, we introduce Projective Radial Distortion Averaging (\prada), a simple yet effective approach for calibrating radially distorted cameras from unordered image collections.
SfM methods typically operate in Euclidean space, recovering 3D reconstructions and poses up to translation, rotation, and scaling. Projective framework extends this to projective space, where reconstructions are defined up to a homography that encapsulates all non-distortion parameters. This simplifies distortion optimization and enables image relationships to be expressed without 3D points or poses at any stage of  the entire algorithm.
We observe that standard image matching algorithms, which are typically designed for pinhole cameras, work well even for distorted images.
For each image, we start off by computing point correspondences with every other image using well-established methods.
For each such pair, we then use a robust correspondence-based minimal solver to obtain an estimate of the radial distortion.
These individual estimates are then finally averaged to produce a single consistent model per camera. \cref{fig:initial_scheme}
An overview of the approach can be found in \cref{fig:overview}.
Our contributions are summarized as follows:  

\begin{figure}
    \centering
    \includegraphics[width=0.44\textwidth]{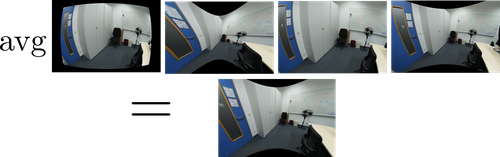}
    \caption{We average multiple imprecise and mutually inconsistent distortion model estimates into a single consistent model.}
    \label{fig:initial_scheme}
\end{figure}

\begin{enumerate}  
    \item We introduce a fully projective method for radial distortion averaging. Our method does not require explicit 3D point reconstruction or focal length estimation, significantly simplifying the auto-calibration process.  
    \item We propose a novel distortion averaging technique that fuses inconsistent pairwise estimates into a single, consistent camera model.  
    \item We demonstrate significant improvements in accuracy and robustness over commonly used methods and validate them on challenging datasets.
\end{enumerate}  

Our method is agnostic to specific image matching techniques and does not rely on keypoint tracks across images, unlike traditional SfM pipelines.

Several methods exist in the literature for estimating radial distortion from a single pair of camera views (\cf \cref{sec:related-works}).
However, two views often lack sufficient information for a fully accurate distortion estimate across the entire image.
The accuracy depends on factors like visual overlap and the quality of pixel correspondences.

\section{Background}
\begin{figure}[t]
    \centering
    \setlength\tabcolsep{0.8pt}
    \begin{tabular}{ccc}
        \rotatebox[origin=c]{90}{\textbf{Distorted}} &
        \begin{minipage}[c]{0.45\columnwidth}
            \centering
            \includegraphics[width=\textwidth]{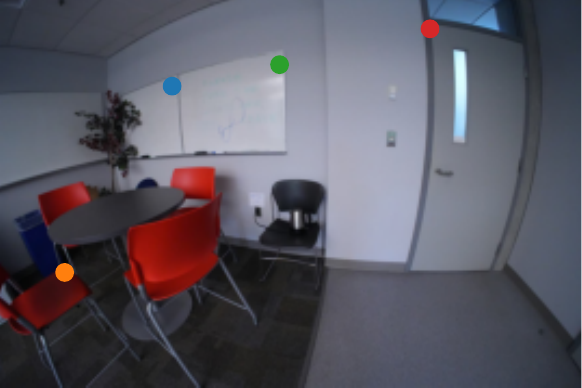}
        \end{minipage} &
        \begin{minipage}[c]{0.45\columnwidth}
            \centering
            \includegraphics[width=\columnwidth]{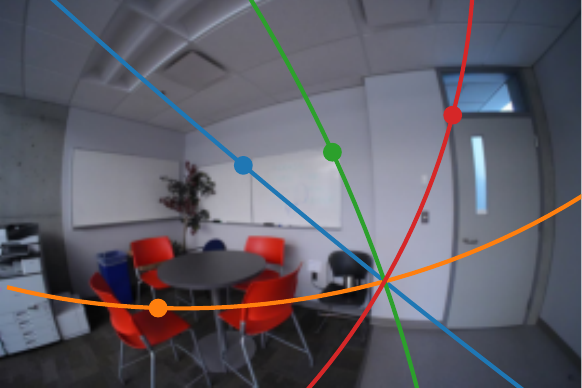}
        \end{minipage} \\[1.1cm]
        \rotatebox[origin=c]{90}{\textbf{Undistorted}} &
        \begin{minipage}[c]{0.45\columnwidth}
            \centering
            \includegraphics[width=\textwidth]{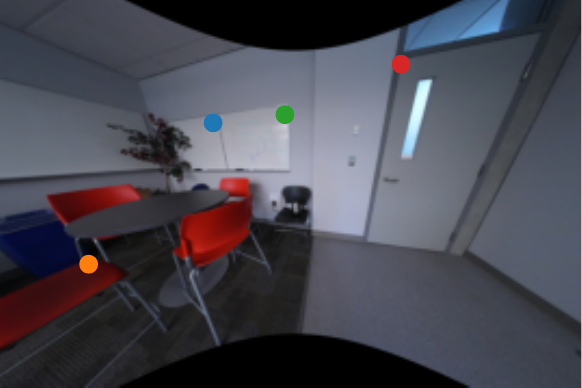}
        \end{minipage} &
        \begin{minipage}[c]{0.45\columnwidth}
            \centering
            \includegraphics[width=\textwidth]{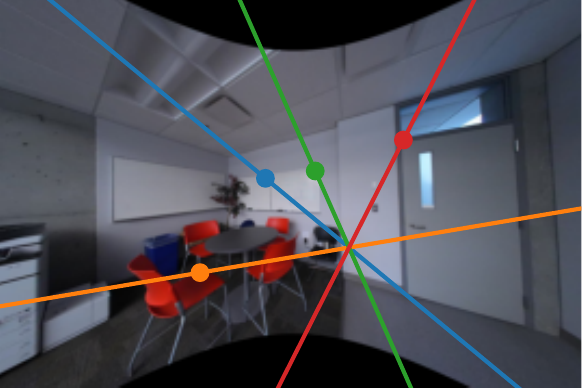}
        \end{minipage}
    \end{tabular}
    \caption{Epipolar lines for the original distorted fisheye image (top), and the undistorted pinhole-like image (bottom). Points in the left and right images indicate matched points. In the undistorted image, epipolar lines are straight, and point correspondences satisfy \cref{eqn:fundamental-constraint}.}
    \label{fig:epipolar_lines}
\end{figure}

\inparagraph{Epipolar geometry with radial distortion:} 
The goal of camera distortion calibration is to map a \textit{distorted} image (\cref{fig:epipolar_lines}, top row) to an \textit{undistorted} image (\cref{fig:epipolar_lines}, bottom row), which is consistent with a pinhole camera.
For pinhole cameras, the epipolar geometry between two views takes a particularly simple form.
Namely, it can be characterized fully by a \(3 \!\times\! 3\) fundamental matrix $F$, which maps points from one image to their corresponding epipolar lines on the other image~\cite{Hartley:2003:MVG:861369}.
Given two pinhole camera images, any 2D point correspondence $\upfirst{i}$, $\upsecond{i}$ will satisfy the \textit{epipolar constraint}:  
\begin{equation}  
\label{eqn:fundamental-constraint}  
    (\upsecondbar{i})^\intercal F \upfirstbar{i} = 0.
\end{equation}  
Here, we use a bar to denote the homogeneous representation of a point \(p\), that is: \(\bar{p} = (p^\intercal; 1)^\intercal\). %

In the presence of distortion, epipolar lines are no longer straight, as shown in the top row of \cref{fig:epipolar_lines}.
This means point correspondences \(\pfirst{i}, \psecond{i}\) will generally not satisfy \cref{eqn:fundamental-constraint} for any single fundamental matrix.
However, there is a mapping \(U_\theta(p)\), with parameters \(\theta\), which maps distorted points \(p\) to undistorted points:
\eq{
\label{eqn:undistorted-points}
    \upfirst{i} = \U{\param}(\pfirst{i}),\quad \upsecond{i} = \U{\param}(\psecond{i}),
}
which are consistent with pinhole geometry.
In particular, we consider \textit{radial distortion}, where the distortion only changes the distance from the center of the image:
\eq{
    U_\theta(p) = d_\theta(\|p\|) p.
}

\inparagraph{Radial distortion calibration:} The goal of radial distortion calibration is formulated as follows: determine the parameters \(\param\) such that the undistorted points \(\upfirst{i}\), \(\upsecond{i}\) satisfy \cref{eqn:fundamental-constraint} for some fundamental fundamental matrix \(F\).
Once the distortion parameters have been estimated, the undistorted points can be used in any 3D reconstruction method designed for pinhole cameras.

\inparagraph{Division camera model:}
Importantly, the epipolar constraint from \cref{eqn:fundamental-constraint} is invariant to multiplication by any non-zero scalar.
We can use this fact to rescale the homogeneous points \( \upfirstbar{i}\) and move the dependence on radial distortion to the last homogeneous coordinate:

\eq{
    \upfirstbar{i} = \begin{pmatrix}
        d_\theta(\|p_i\|)p_i
    \\  1
    \end{pmatrix} \simeq\begin{pmatrix}
        p_i
    \\  1 / d_\theta(\|p_i\|)
    \end{pmatrix}
}
Where \(\simeq\) denotes equivalence up to a scaling factor.
Of particular importance to our work is the one-parameter division camera model~\cite{fitzgibbon2001simultaneous}, given by: $d_\lambda(r) = 1/(1 + \lambda r^2)$.
Which has the undistortion function:
\eq{
    U_{\lambda}(\bar{\pfirst{i}}) \simeq \begin{pmatrix}
        \pfirst{i} \\ 1 + \lambda \|\pfirst{i}\|^2
    \end{pmatrix}.
    \label{eq:division_model_first}
}
While the one-parameter division model has well-established minimal solvers \cite{fitzgibbon2001simultaneous,kukelova2008F9A,Kukelova2015F10}, it is not expressive enough to capture many real-world distortion patterns. Therefore, we also consider the extension to a k-th degree polynomial~\cite{larsson2019revisiting2d3dRadial}:
\eq{
    U_\theta(\bar{\pfirst{i}}) = \begin{pmatrix}
        \pfirst{i} \\ h_\theta(\|\pfirst{i}\|)
    \end{pmatrix}
    \label{eq:division-model-multiple}
}
where $h(r)=\sum_{i=0}^k\theta_ir^i$.
Which corresponds to $d(r) =  1/h_\theta(r)$.

%% file: sec/2_related_work.tex
\begin{figure*}
    \centering
    \def\svgwidth{\linewidth}
    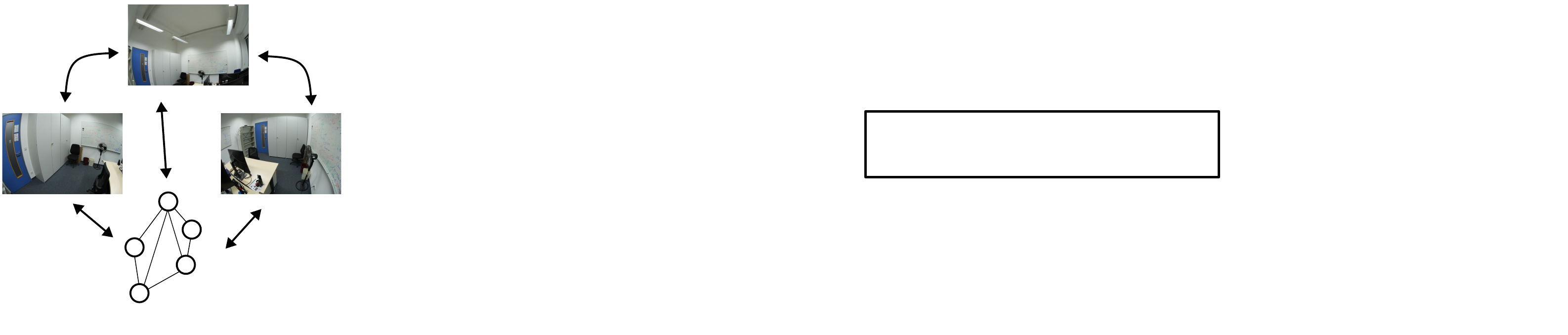
    \caption{\textbf{Method overview:} Each possible image pair generates separate parameters of the distortion model via a robust solver (\cf \cref{sec:robust-solver}). Then, for each camera $C_i$, the models are merged to form a single consistent estimation $\theta_i$ using \textit{distortion averaging} (\cf \cref{sec:distortion_averaging}). Each stage is further refined using nonlinear optimization (\cf \cref{sec:nonlinear-optimization,sec:global_pa})}. %
    \label{fig:overview}
\end{figure*}

\section{Related work}
\label{sec:related-works}
Accurate distortion estimation plays a crucial role in tasks such as SfM and 3D reconstruction. Various approaches have been proposed to address this challenge, each with different assumptions and requirements.

One category of approaches leverages a known 3D map to estimate camera distortions and focal lengths. These methods can be divided into two subcategories. The first assumes a perfect 3D map, which is the standard setup for camera calibration using patterns like chessboards~\cite{zhang2000flexible,scaramuzza2006toolbox,mei2007single} or learned patterns~\cite{xian2023neural}. The second relies on a 3D map constructed by a SfM framework to register a new image within it.
For an available 3D map, a 3-parameter polynomial distortion model can be obtained from 5 2D-to-3D correspondences \cite{kukelova20132d3drealtime}.
If more correspondences are available, calibration can be estimated using a  rational model \cite{larsson2019revisiting2d3dRadial} or implicitly \cite{pan2022implicitDistortion}.
However, obtaining an accurate 3D map from images with unknown distortion can be a challenging problem.

In the case when a 3D map is not available, existing approaches can be broadly split into three categories: N-point-solvers, bundle-adjustment-based, and learning-based.

\inparagraph{N-point solvers} work by solving for the unknown fundamental matrix and distortion parameters subject to the constraints from \cref{eqn:undistorted-points} and \cref{eqn:fundamental-constraint}.
Pioneering work from \citet{fitzgibbon2001simultaneous} shows that a single-parameter polynomial model can be estimated minimally using 8 correspondences from an image pair from the same camera.
The method was extended by \citet{jiang2015minimal} to include focal length estimation.
If the two images come from different cameras, a one-parameter (per camera) polynomial model can be estimated minimally with 9 correspondences \cite{Byrod2008F9, kukelova2008F9A}, by 3 variants of solvers requiring 10 correspondences \cite{Kukelova2015F10}, with 12 correspondences \cite{Byrod2008F9} or 15 correspondences \cite{barreto2005F15}.
SVA \cite{lochman2021minimal} goes beyond point correspondences and detects vanishing points by finding the intersection of circular arcs (corresponding to distorted lines). However, because of this reliance on straight lines, it may not work in cases where straight lines are not present.
According to performance analysis in \cite{Kukelova2015F10}, the 10-point method F10 gives the best result in terms of performance compared to minimal methods.

\inparagraph{Bundle-adjustment methods}
Distortion parameters can also be calculated as part of a SfM pipeline in the bundle adjustment step \cite{schoenberger2016sfm, schoenberger2016mvs, pan2024glomap, hagemann2023droidCalib}, or NeRF optimization \cite{Jeong2021selfcalibrating}, although at the cost of increasing the complexity of the loss function. Global SfM methods such as GLOMAP~\cite{pan2024glomap} are promising, as they utilize least squares fitting for each pair of images and then average the results, potentially allowing distortion estimation in the first stage.
Calibration-free SfM \cite{larsson2020calibration} 
 can avoid the need to estimate radial distortion. This approach uses clever minimal solvers that require three images in the dataset to have intersecting optical axes or tracks on 4 images \cite{hurby2023fourview}, which may not always be the case, especially for sparse images.
RpOSE \cite{iglesias2021distortionInvariant} solves the projective reconstruction globally in a radial distortion-invariant way to later fit radial distortion. Having a big basin of convergence, this approach requires a set of image points tracked along several images, which can be challenging to obtain without pre-filtering stages that should be carefully handled in the presence of distortion.

\inparagraph{Learning-based methods}
Calibration parameters can also be estimated using learning-based methods. 
Early works model distortion by learning to map point correspondences directly to 3d points \cite{mendoncca2002neuralNetworks} or by learning the residual error induced by the unknown calibration \cite{wen1991hybrid, do1999application}.
\citet{rong2017CNNradial} predicts a single-parameter division model. Trained on synthetically distorted images. %
DeepFocal \cite{workman2015deepfocal} predicts focal length, trained on internet images.
\citet{hold2018perceptual} Predicts focal length and rotation relative to the horizon line.
DeepCalib~\cite{bogdan2018deepcalib} predicts the focal length and $\xi$ from the unified spherical model \cite{geyer2000unifyingpanoramic, barreto2006unifyingCentral, mei2007single}, trained on crops from panoramic images.
DroidCalib \cite{hagemann2023droidCalib} extents DROID-SLAM~\cite{teed2021droid} to optimize unified camera model~\cite{mei2007single} with Gauss-Newton. 
GeoCalib \cite{veicht2024geocalib} regresses a pixel-wise perspective field \cite{jin2023perspective}, which can be used to estimate radial distortion with nonlinear optimization. In DeepCalib~\cite{bogdan2018deepcalib}, the authors observe that their solution performs better on images with significant distortion. They explain this by pointing to the non-uniformity of the training data, noting the absence of a dataset containing diverse fisheye distortions. Additionally, they argue that manually generating such data is unfeasible because applying artificial fisheye distortion to an image produces unrealistic results.

Similarly to \citet{sarlin2021back}, we argue that re-learning geometric principles with neural networks is often unnecessary.
Instead, these principles can be accurately handled with traditional methods, using deep learning where exact modeling is infeasible.  
We find that well-established learning-based matchers  ~\cite{lowe2004distinctive,sarlin2020superglue,lindenberger2023lightglue,sun2021loftr,detone2018superpoint} demonstrate strong robustness to distortion changes~\cite{Kukelova2015F10,kukelova20132d3drealtime,larsson2019revisiting2d3dRadial,pan2022implicitDistortion,kukelova2008F9A,schoenberger2016mvs,schoenberger2016sfm,pan2024glomap}.

\inparagraph{Eucledian vs Projective SfM}
In SfM, reconstruction can be performed in either a projective \cite{kasten2019gpsfm,kasten2019algebraic} or Euclidean \cite{pan2024glomap,schoenberger2016sfm} framework \cite{Hartley:2003:MVG:861369}. Projective SfM recovers the scene structure up to a homography without requiring information on focal length. The resulting reconstruction can still be optimal regarding reprojection errors and geometric relations but lacks metric accuracy. The point cloud can be arbitrarily skewed. Often, projective reconstruction is utilized as a first step in SfM methods \cite{iglesias2021distortionInvariant,iglesias2023expose,weber2024power} because it requires estimating fewer parameters. \cite{kasten2019gpsfm} is particularly interesting because it does not depend on tracking points across several images.

%% file: figures/scheme.pdf_tex
\begingroup%
  \makeatletter%
  \providecommand\color[2][]{%
    \errmessage{(Inkscape) Color is used for the text in Inkscape, but the package 'color.sty' is not loaded}%
    \renewcommand\color[2][]{}%
  }%
  \providecommand\transparent[1]{%
    \errmessage{(Inkscape) Transparency is used (non-zero) for the text in Inkscape, but the package 'transparent.sty' is not loaded}%
    \renewcommand\transparent[1]{}%
  }%
  \providecommand\rotatebox[2]{#2}%
  \newcommand*\fsize{\dimexpr\f@size pt\relax}%
  \newcommand*\lineheight[1]{\fontsize{\fsize}{#1\fsize}\selectfont}%
  \ifx\svgwidth\undefined%
    \setlength{\unitlength}{1522.5769043bp}%
    \ifx\svgscale\undefined%
      \relax%
    \else%
      \setlength{\unitlength}{\unitlength * \real{\svgscale}}%
    \fi%
  \else%
    \setlength{\unitlength}{\svgwidth}%
  \fi%
  \global\let\svgwidth\undefined%
  \global\let\svgscale\undefined%
  \makeatother%
  \begin{picture}(1,0.21370406)%
    \lineheight{1}%
    \setlength\tabcolsep{0pt}%
    \put(0,0){\includegraphics[width=\unitlength,page=1]{figures/scheme.pdf}}%
    \put(0.03937607,0.00315254){\color[rgb]{0,0,0}\makebox(0,0)[lt]{\lineheight{1.25}\smash{\begin{tabular}[t]{l}Image graph matching\end{tabular}}}}%
    \put(0,0){\includegraphics[width=\unitlength,page=2]{figures/scheme.pdf}}%
    \put(0.29187556,0.03813702){\color[rgb]{0,0,0}\makebox(0,0)[lt]{\lineheight{1.25}\smash{\begin{tabular}[t]{l}Initial Pairwise\\Estimations\end{tabular}}}}%
    \put(0.52836036,0.02482233){\color[rgb]{0,0,0}\makebox(0,0)[lt]{\lineheight{1.25}\smash{\begin{tabular}[t]{l}Distortion Averaging\\\end{tabular}}}}%
    \put(0.52011395,0.13512131){\color[rgb]{0,0,0}\makebox(0,0)[lt]{\lineheight{1.25}\smash{\begin{tabular}[t]{l}...\end{tabular}}}}%
    \put(0,0){\includegraphics[width=\unitlength,page=3]{figures/scheme.pdf}}%
    \put(0.80131918,0.02172497){\color[rgb]{0,0,0}\makebox(0,0)[lt]{\lineheight{1.25}\smash{\begin{tabular}[t]{l}Global Projective BA\end{tabular}}}}%
    \put(0,0){\includegraphics[width=\unitlength,page=4]{figures/scheme.pdf}}%
  \end{picture}%
\endgroup%

%% file: sec/3_method.tex
\section{Method}
\label{sec:method}

We start by estimating 2D-2D correspondences $\left( \pfirst{i}, \psecond{i}\right)_{mn}$ for each pair of images $\{I_m, I_n\}$. These correspondences are the basis for estimating the distortion parameters of the camera model for each image.
From each correspondence set $\left( \pfirst{i}, \psecond{i}\right)_{mn}$, we then estimate a fundamental matrix and one-parameter distortion model (\cref{sec:robust-solver}). This results in multiple initial estimates for each image's camera model. These distortion estimates are further refined by minimizing the Sampson error \cite{rydell2024revisiting, harker2006first} (\cref{sec:nonlinear-optimization}).

The two-view distortion estimates are typically accurate only in the regions of the image where 2D points are available.
So, to obtain a reliable camera model across the full image, we fuse the individual estimates through \textit{distortion averaging} (\cref{sec:distortion_averaging}). This step ensures a single consistent camera model per camera, which \textit{optimally} merges the individual models.

As a final step to further improve the distortion estimate, we perform a global refinement (\cref{sec:global_pa}) of the Sampson error \cite{rydell2024revisiting, harker2006first} across all images.
The full pipeline, illustrated in \cref{fig:overview}, produces a highly reliable distortion estimate without estimation of 3D points or camera poses.

\subsection{Two-view initialization}
\label{sec:robust-solver}
We obtain initial one-parameter distortion models using LO-RANSAC \cite{chum2003locally} with the F10 solver proposed by \citet{Kukelova2015F10}.
This minimal solver supports different camera models for both images.
For each image pair $\{I_m, I_n\}$, we have a separate estimation of 2 distortion models.
We normalize pixel coordinates by the length of the image diagonal to increase the numerical stability of the solver.

The benefits of this stage are two-fold.
First off, we obtain initial estimates of the distortion models and fundamental matrices for all cameras from all matched image pairs, respectively.
Secondly, RANSAC ensures that outlier correspondences generated by the matcher are filtered out.

\subsection{Two-view nonlinear refinement}
\label{sec:nonlinear-optimization}
Next, we refine the initial single-parameter distortion estimates using nonlinear refinement with a higher-order degree-$k$ polynomial model from \cref{eq:division-model-multiple}.
Following \citet{scaramuzza2006toolbox}, we parameterize polynomials such that $\theta_0$ is fixed to 1 and $\theta_1$ is fixed to 0.
This parameterization ensures the model behaves like a pinhole near the image center.
Each higher-order model is initialized by setting $\theta_0 = 1, \theta_2 = \lambda$, and other $\theta_i$ to 0.

Minimal solvers for the higher-order polynomial division model are impractical due to the increasing number of minimal samples required. Therefore, we rely on nonlinear optimization for this stage.
For this we use the Sampson error~\cite{rydell2024revisiting,harker2006first}:
\eq{
\label{eq:sampson-error}
r^2_{\text{sampson}} \left(p, q, F, \theta_1, \theta_2\right) = %
\frac{C^2(dp, dq)}{\left\lVert J^C_{dp} \right\rVert^2 + \left\lVert J^C_{dq} \right\rVert^2},
}
where
\eq{ C(dp, dq) = \U{\param_1}(q + d q)^\intercal F \U{\param_2}(p + d p) }
is the epipolar constraint centered at $p, q$.
Geometrically, the Sampson error approximates the minimum adjustment in pixels required for each point correspondence $\pfirst{i}, \psecond{i}$ to satisfy the epipolar constraint with respect to $F$:
\eq{
    r^2_{\text{sampson}} \approx \begin{cases}
    \min\limits_{dp, dq} \lVert dp \rVert^2_2 + \lVert dq \rVert^2_2 \\
    \text{s.t.} \quad\!\! \U{\param_1}(q_i \!+\! dq)^\intercal F \U{\param_2}(p_i \!+\! dp) = 0.
    \end{cases}
    \label{eqn:min-fundamental-adjustment}
}
The approximation is obtained by linearizing the epipolar constraint around $p_i, q_i$.

Let $(I_i, I_j)$ be an image pair with $m$ correspondences. Let $C_i$ and $C_j$ be the corresponding cameras.
The camera parameters estimated during the two-view step (\cref{sec:robust-solver}) are denoted as $\theta_{C_i}^{ij}$ and $\theta_{C_j}^{ij}$. 
The superscript indicates that the parameters are specific to this image pair.
The optimization problem can then be formulated as:
\eq{
\underset{F_{ij}, \theta_{C_i}^{ij}, \theta_{C_j}^{ij}}{\operatorname{argmin}} \sum\limits_{l=1}^m r^2_{\text{sampson}} \left( p_l, q_l, F_{ij}, \theta_{C_i}^{ij}, \theta_{C_j}^{ij} \right)
\label{eqn:refinement}
}
Intuitively, minimizing the Sampson error refines the camera parameters and the fundamental matrices to ensure that the adjustments needed for the points to satisfy the epipolar constraint \cref{eqn:fundamental-constraint} are as small as possible.
In addition, as \cref{eqn:refinement} is defined only in terms of the epipolar constraint, it lets us bypass the explicit estimation of 3D points.

Nonlinear optimization of the fundamental matrix requires careful handling. While being $3 \times 3$ real matrix it has 7 degrees-of-freedom~\cite{bartoli2004nonlinear}. One common way to parameterize it is $SO(3) \times S^1 \times SO(3)$. 
 $SO(3)$ is the group of $3 \times 3$ rotation matrices in $\mathbb{R}^3$ and has 3 degrees of freedom. $S^1$ is the 1-dimensional sphere representing the unit circle and can be parameterized as an angle. This parameterization is unique and supports local updates via the corresponding Lie group exponential map \cite{bartoli2004nonlinear}

The fundamental matrix is scale-invariant, i.e. it remains unchanged when multiplied by any non-zero scalar, and it has at most two non-zero singular values \cite{Hartley:2003:MVG:861369}.
Given these properties, the minimal parameterization of the fundamental matrix can be obtained using the following scheme:

\begin{enumerate}
    \item Compute SVD $F = U \Sigma V^\top$.
    \item Normalize $U$ and $V$ to ensure they have determinant 1:
    \eq{
        U &\leftarrow U \cdot \operatorname{det}(U), \\
        V &\leftarrow V \cdot \operatorname{det}(V).
    }
    \item Normalize singular values $\sigma_1$ and $\sigma_2$:
    \eq{
    \sigma' = \frac{(\sigma_1, \sigma_2)}{\|\sigma_1, \sigma_2\|_2}.
    }
\end{enumerate}
After these steps $U \in SO(3), V \in SO(3), \sigma' \in S^1$.

\subsection{Polynomial distortion regularization}
\label{sec:polynomial_regularization}
Since the images are not fully covered by 2D-2D correspondences $\left( \pfirst{i}, \psecond{i}\right)_{mn}$, the camera model is unconstrained and can behave arbitrarily in the uncovered regions. 
To address this, we introduce additional regularization to the distortion polynomials.

Following the approach of \citet{pan2022implicitDistortion, hartley2007parameter} we apply the \textit{local linearity assumption} to the undistortion function, treating it as a function of the radius.
This aims to constrain the rate of change of the undistortion function, ensuring it remains monotonic. In the continuous case, this regularization can be formulated as:
\eq{ \min \int\limits_0^R \left\lVert \frac{d U_\theta(r)}{dr} \right\rVert^2 dr}
This function is not analytically integrable due to the potentially high degree polynomial in the denominator.
Instead, we use an efficient numerical approximation of this integral.

\subsection{Distortion averaging}
\label{sec:distortion_averaging}
After the initial pairwise estimation of camera parameters, we collect all the observed parameters for each physical camera model. The goal of this step is to determine the model that best fits the set of estimated camera parameters.
For each camera $C$, we have multiple, potentially mutually inconsistent estimates of the distortion parameters $\theta$. %
Each estimated distortion model is generally consistent only within the regions covered by point correspondences and may be inconsistent in other image regions.
To combine these individual estimates into a single, unified distortion polynomial, we propose \textbf{distortion averaging}. We solve an optimization problem in functional space so that the solution behaves similarly to each of the estimated camera models within the image.

As a general principle, we can average multiple distortion models $\hat U_i$, $i=1,\ldots,n$ by solving a weighted least-squares problem within the space of functions on the image plane:
\eq{
    \label{eqn:distortion-averaging}
    \bar U &= \underset{U}{\operatorname{argmin}}\sum_{i=1}^n \omega_i \left\langle U, \quad\!\!\!\!\hat U_{i} \right\rangle^2 \\
     &= \underset{U}{\operatorname{argmin}}\sum_{i=1}^n \omega_i \int_{p\in I} \left\lVert U(p) - \hat U_{i}(p) \right\rVert^2\diff{x}\diff{y} \\
}
where $p=(x, y)$ ranges over all pixels of the image, and $\omega_i$ are weights that sum to 1.
Reparametrizing in terms of the radial distortion model $U(p) = pd(\|p\|)$ and expressing this integral in polar coordinates, we get:
\eq{
\label{eq:integral_polar}
    \bar d  &= \underset{d}{\operatorname{argmin}}\sum_{i=1}^n \omega_i \int\limits_0^R \left\lVert d(r) - \hat d_{i}(r) \right\rVert^2 r^3 \diff{r}.
}
Finally, we constrain the solution to be in the form of an undistortion function of the polynomial division model (See \cref{eq:division-model-multiple}). Then \cref{eq:integral_polar} changes to:
\eq{
\label{eq:distortion_averaging}
    \bar \theta  &= \underset{\theta}{\operatorname{argmin}}\sum_{i=1}^n \omega_i \int\limits_0^R \left\lVert \frac{1}{h_\theta(r)} - \frac{1}{h_{\theta_i}(r)} \right\rVert^2 r^3 \diff{r}
}
where $\theta$ corresponds to the coefficients of the polynomial. We solve this numerically, initializing $\theta$ as a weighted average of $\theta_i$:
\eq{
\label{eq:distortion_averaging_dumm}
    \bar\theta = \frac{\sum_{i=1}^n \omega_i\theta_i}{\sum_{i=1}^n \omega_i}.
}
Note that our averaging approach can generate polynomials of nearly any degree, making the degree a hyperparameter of the averaging scheme. However, the degree of $\theta$ from \cref{eq:distortion_averaging_dumm} is limited by the maximum degree of $\theta$.

While \cref{eq:distortion_averaging_dumm} is not in general optimal for \cref{eq:distortion_averaging}, it turns out to be an optimal solution of \cref{eq:integral_polar} for the multiplicative distortion model~\cite{larsson2019revisiting2d3dRadial} (see Supplementary material).

\subsection{Global refinement}
\label{sec:global_pa}
Even after the distortion averaging step, the resulting camera models may still lack sufficient accuracy. We refine them together with the fundamental matrix in a global optimization.
We minimize a robust Sampson error loss across all images:
\eq{
\label{eq:sampson_full}
\underset{\{F_{ij}\}, \{\theta_k\}}{\operatorname{argmin}}  \! \sum\limits_{l, i, j} \rho \left(r_{\text{sampson}} \left( p_l, q_l, F_{ij}, \theta_{C_i}, \theta_{C_j} \right) \right)
}
where $\rho$ is a Cauchy loss function \cite{mlotshwa2022cauchy} and $p_l, q_l$ corresponding points.
If an image $I_i$ appears in multiple pairs, such as $(I_i, I_j)$ and $(I_i, I_k)$, the same camera model is used for all pairs, and the optimization is performed jointly over these pairs.
Solving \cref{eq:sampson_full} ensures that the camera models remain consistent with the relative geometry across multiple image pairs. In addition, this loss formulation allows the optimization of principal points if there is sufficient coverage across all image pairs.

%% file: sec/X_suppl.tex
\clearpage
\maketitlesupplementary
\appendix

\section{Implementation Details}
\label{sec:supp_implementation_details}
Following best practices from COLMAP, we estimate initial pairwise models in parallel with image matching. This adds almost no overhead to the feature-matching process. Our method then starts with already estimated pairwise models.
We implement the proposed method in C++. Unlike in incremental reconstruction, the global approach allows parallel computation of all components, providing true scalability, so we parallelize every possible step. We use TBB \cite{10.5555/1352079.1352134} for parallel execution on the CPU. Since all derivatives are computed in closed form, a CUDA implementation is also feasible but beyond the scope of the current work.

For the optimization phase of LO-RANSAC \cite{chum2003locally}, we compute the derivatives of the Sampson error with respect to both cameras and fundamental matrix according to the parameterization described in \cref{sec:nonlinear-optimization}. Our experiments show that, typically, ten iterations of local refinement are sufficient. The weights in \cref{eq:distortion_averaging} are based on the area covered by the matches. This way, the actual overlap between images is taken into account compared to the naive number of inliers. This handles outlier cameras that may occur after pairwise estimation. The final global refinement (see \cref{sec:global_pa}) is done with analytical derivatives using nonlinear solvers from ceres-solver~\cite{Agarwal_Ceres_Solver_2022}.
We run it several times, re-estimating the inliers between successive runs. In the first pass of the global projective refinement, we fix the camera centers and optimize only the distortions and fundamental matrices. The camera centers are then optimized while all other camera parameters are fixed.

\section{Performance by sequence length}
\begin{figure}
    \centering
    \includegraphics[width=\columnwidth]{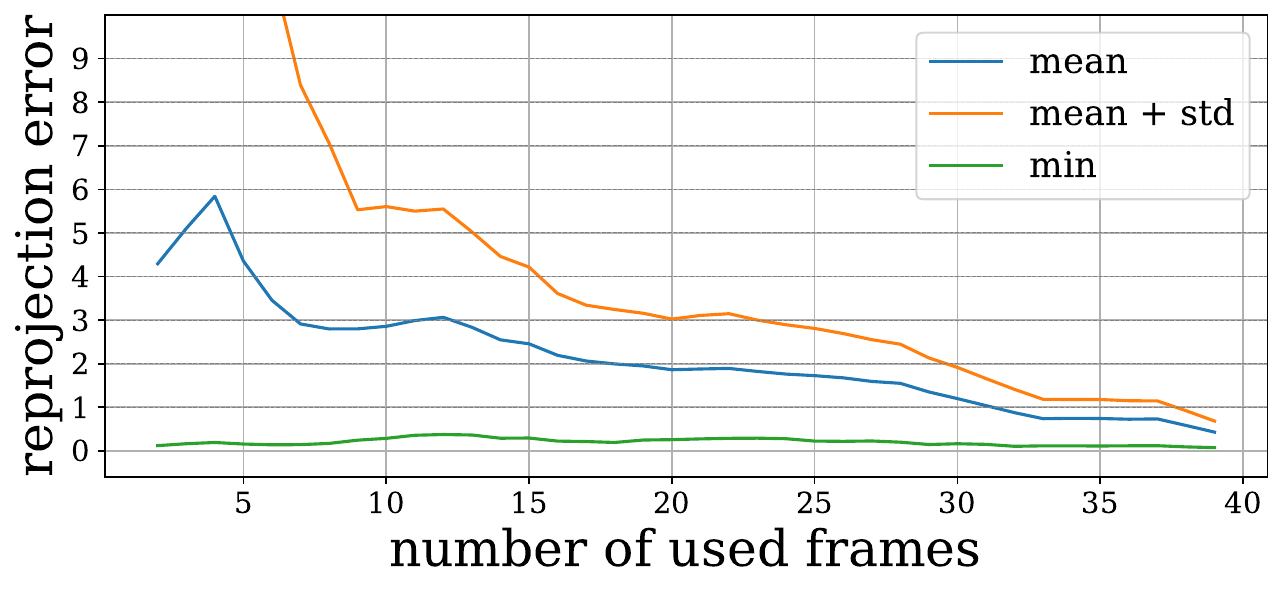}
    \caption{\textbf{Mean error for randomly sampled N-frame sequences:} We study the behavior of the proposed algorithm with different numbers of input frames. We sample 20 random sequences of frames for each \textit{number of frames} configuration. We find that 31 frames are sufficient to obtain a competitive 1px mean error.}
    \label{fig:scanet_hist_multi}
\end{figure}
We evaluate the performance of the proposed algorithm on the ScanNet++ dataset under varying numbers of input frames.
For each sequence length $N$, ranging from 2 to 40, we randomly sample 1000 $N$-frame subsequences (20 from each of the 50 ScanNet++ test sequences).
We then run our method on each subsequence and plot the average focal-adjusted reprojection error in \cref{fig:scanet_hist_multi}.
Frames are 5 frames apart, relative to the original ScanNet++ frame rate.%

Comparing with the full-sequence results in \cref{tab:projection_errors}, we observe that with 31 input frames, the proposed algorithm achieves a mean error (1.0 px) lower than that of Colmap~\cite{schoenberger2016mvs,schoenberger2016sfm} (2.0 px) and Glomap~\cite{pan2024glomap} (1.8 px), even when these methods process the entire frameset.

\section{Distortion averaging for the multiplicative model}
Our radial distortion averaging formulation can be adapted for any distortion parametrization $d_\theta(r)$. 
In terms of the model parameters, $\theta$, the associated optimization problem (repeated here for convenience) is formulated as:
\eq{
    \label{eq:parameter-averaging}
    \bar \theta  &= \underset{\theta}{\operatorname{argmin}}\sum_{i=1}^n \omega_i \int\limits_0^R \left\lVert d_\theta(r) - d_{\theta_i}(r) \right\rVert^2 r^3 \diff{r}.
}
For our method, we use the divisional distortion model: $d_\theta(r) = 1/h_\theta(r)$, where $h_\theta(r)=\sum_{j=0}^k\theta_jr^j$ is a degree-$k$ polynomial with coefficients given by the vector $\theta$.
Primarily because of the availability of the well-proven F10 solver \cite{Kukelova2015F10} and computationally efficient derivatives for the Sampson error~\cite{rydell2024revisiting,harker2006first}.

For the division model, \cref{eq:parameter-averaging} cannot be evaluated exactly, so we optimize over a numerical discretization with uniform spacing.
Interestingly, it turns out that for the \textit{multiplicative distortion model}, the average distortion as defined by \cref{eq:parameter-averaging} corresponds to averaging the distortion parameters.
That is:
\eq{
    \label{eq:weighted-average-parameter}
    \bar\theta = \frac{\sum_{i=1}^kw_i\theta_i}{\sum_{i=1}^kw_i}
}
when $d_\theta(r)$ is parametrized as a degree-$k$ polynomial: $d_\theta(r) = h_\theta(r)$.
To see this, let us rewrite \cref{eq:parameter-averaging} for the multiplicative distortion model: %
\eq{
    \label{eq:parameter-averaging-multiplicative}
   \bar \theta  &= \underset{\theta}{\operatorname{argmin}}\underbrace{\sum_{i=1}^n \omega_i \int\limits_0^R \left\lVert \sum_{j=0}^k\theta_jr^j - \sum_{j=0}^k\theta^i_jr^j \right\rVert^2 r^3 \diff{r}}_{:= L(\theta)}
}
To find the minimum, we take the derivatives of $L(\theta)$ with respect to the coefficients $\theta_t$:
\eq{
    \label{eq:multiplicative-derivative}
    \frac{dL}{d\theta_t} &=
    2\sum_{i=1}^n \omega_i \int\limits_0^R \left( \sum_{j=0}^k\theta_jr^j - \sum_{j=0}^k\theta^i_jr^j \right) r^{3+t} \diff{r} \\
    &= 2 \sum_{i=1}^n\sum_{j=0}^kw_iA_{tj}(\theta_j - \theta_j^i).
}
Where 
\eq{
    A_{ti} = \int\limits_0^R r^{j+t+3} \diff{r} = \frac{R^{j+t+4}}{j+t+4}.
}
Rearranging and solving for $\frac{dL}{d\theta_t} = 0$ we get:
\eq{
    \sum_{j=0}^kA_{tj}\theta_j = \frac{1}{\sum_{i=1}^n \omega_i}\sum_{j=0}^kA_{tj}\sum_{i=1}^n \omega_i\theta_j^i
}
which in matrix form corresponds to
\eq{
    A\theta = A \bar\theta
}
where $\bar\theta$ is the weighted average parameter, as defined in \cref{eq:weighted-average-parameter}.
Since $A$ is invertible, we get that the solution to \cref{eq:parameter-averaging-multiplicative} is $\theta = \bar\theta$.
In this case, $\theta$ is independent of the radius $R$. This means that for \textit{multiplicative distortion model}, averaging across an image is equivalent to averaging across the entire space of $\mathbb{R}^2$, including areas outside the image.